\documentclass{article}

\usepackage{microtype}
\usepackage{graphicx}
\usepackage{subfigure}
\usepackage{booktabs} % for professional tables
\usepackage{amsfonts}
\usepackage{amsmath}
\usepackage{amssymb}
\usepackage{multirow}
\usepackage{hyperref}
 \usepackage{ulem}
\usepackage[accepted]{icml2021}

\icmltitlerunning{Meta-Learned Invariant Risk Minimization}

\begin{document}

\twocolumn[
\icmltitle{Meta-Learned Invariant Risk Minimization}

% It is OKAY to include author information, even for blind
% submissions: the style file will automatically remove it for you
% unless you've provided the [accepted] option to the icml2021
% package.

% List of affiliations: The first argument should be a (short)
% identifier you will use later to specify author affiliations
% Academic affiliations should list Department, University, City, Region, Country
% Industry affiliations should list Company, City, Region, Country

% You can specify symbols, otherwise they are numbered in order.
% Ideally, you should not use this facility. Affiliations will be numbered
% in order of appearance and this is the preferred way.
\icmlsetsymbol{equal}{*}

\begin{icmlauthorlist}
\icmlauthor{Jun-Hyun Bae}{knu}
\icmlauthor{Inchul Choi}{knu}
\icmlauthor{Minho Lee}{knu}
\end{icmlauthorlist}

\icmlaffiliation{knu}{Graduate School of Artificial Intelligence, Kyungpook National University, Daegu, Republic of Korea}

\icmlcorrespondingauthor{Minho Lee}{mholee@gmail.com}

% You may provide any keywords that you
% find helpful for describing your paper; these are used to populate
% the "keywords" metadata in the PDF but will not be shown in the document
\icmlkeywords{Machine Learning, ICML}

\vskip 0.3in
]

% this must go after the closing bracket ] following \twocolumn[ ...

% This command actually creates the footnote in the first column
% listing the affiliations and the copyright notice.
% The command takes one argument, which is text to display at the start of the footnote.
% The \icmlEqualContribution command is standard text for equal contribution.
% Remove it (just {}) if you do not need this facility.

\printAffiliationsAndNotice{}  % leave blank if no need to mention equal contribution
%\printAffiliationsAndNotice{\icmlEqualContribution} % otherwise use the standard text.

\begin{abstract}
Empirical Risk Minimization (ERM) based machine learning algorithms have suffered from weak generalization performance on data obtained from out-of-distribution (OOD).
To address this problem, Invariant Risk Minimization (IRM) objective was suggested to find invariant optimal predictor which is less affected by the changes in data distribution.
However, even with such progress, IRMv1, the practical formulation of IRM, still shows performance degradation when there are not enough training data, and even fails to generalize to OOD, if the number of spurious correlations is larger than the number of environments.
In this paper, to address such problems, we propose a novel meta-learning based approach for IRM.
In this method, we do not assume the linearity of classifier for the ease of optimization, and solve ideal bi-level IRM objective with Model-Agnostic Meta-Learning (MAML) framework. 
Our method is more robust to the data with spurious correlations and can provide an invariant optimal classifier even when data from each distribution are scarce.
In experiments, we demonstrate that our algorithm not only has better OOD generalization performance than IRMv1 and all IRM variants, but also addresses the weakness of IRMv1 with improved stability.
\end{abstract}

\section{Introduction}
\label{introduction}

\begin{figure}[ht]
\vskip 0.2in
\begin{center}
\centerline{\includegraphics[width=0.85\columnwidth]{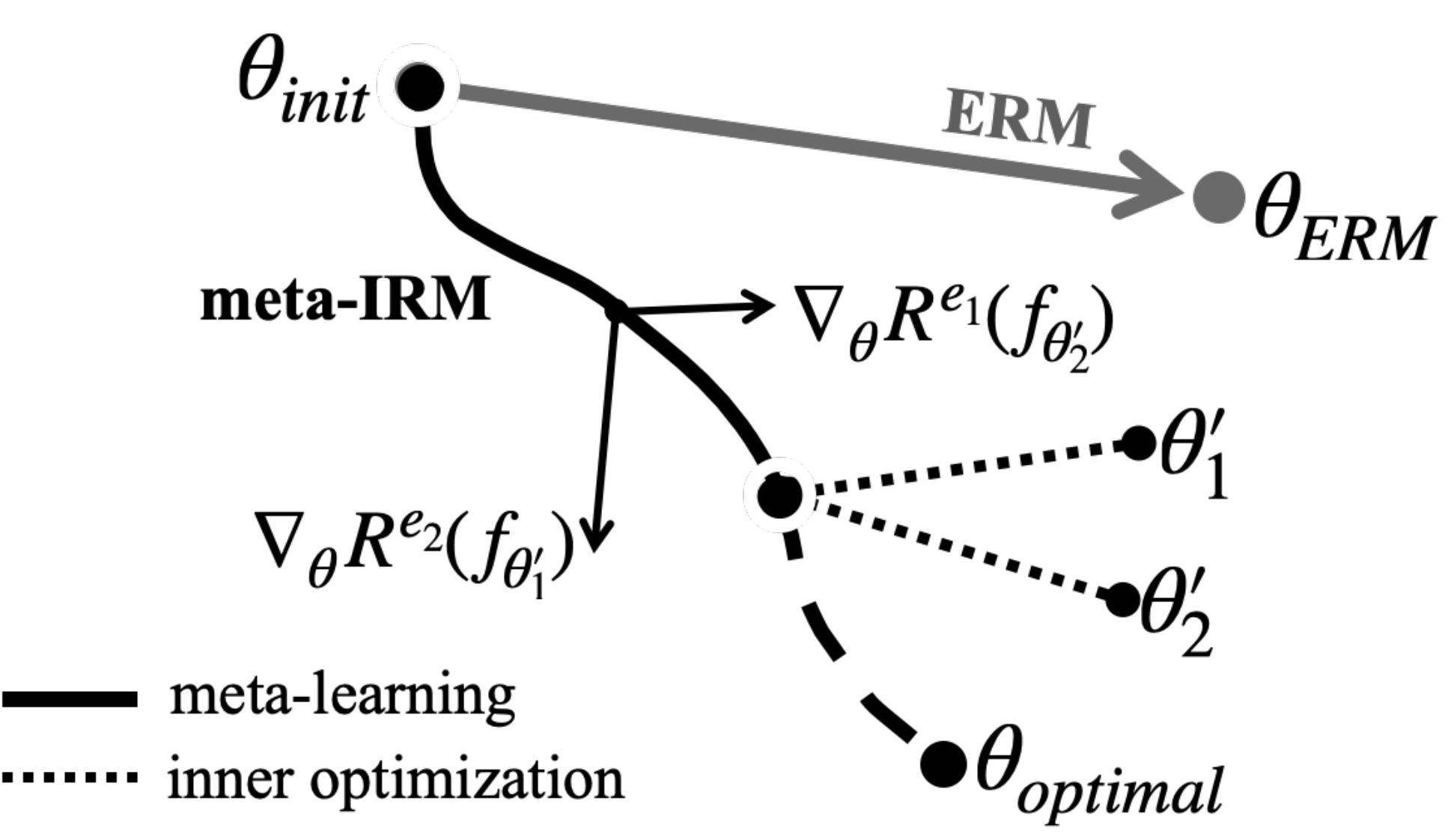}}
\caption{Diagram of our proposed meta-IRM, which learns invariant optimal predictor across the data distribution with meta-learning.}
\label{grad.png}
\end{center}
\vskip -0.4in
\end{figure}

Conventional machine learning algorithms often fail to generalize, when training data include spurious correlations with strong correlations to a target label.
With such data, they are more likely to learn spurious correlations, which does not correspond to the true causal relation that exists in data.
Learning by minimizing empirical risks of data is reasonable choice for the generalization performance of model, if test data are sufficiently similar to a training distribution.
However, if test data include strong spurious correlations, or sampled from the out-of-distribution (OOD), ERM based approach inevitably fails to provide reliable outcomes.
This problem is known as OOD generalization problem and can be easily found in the real-world scenario for machine learning.~\cite{beery2018recognition, ilyas2019adversarial, geirhos2018imagenet, de2019causal}

Invariant Risk Minimization (IRM)~\cite{arjovsky2019invariant} is a recently proposed learning paradigm to address OOD generalization problem with the theory of Invariant Causal Prediction~\cite{peters2016causal, heinze2018invariant}. 
It is based on the fundamental idea that true causal relations remain invariant while spurious correlations can vary across data distributions from different environments.
Based on this idea, \citet{arjovsky2019invariant} introduce IRM objective which finds an invariant predictor that is also optimal for every available data environments. 
In its formulation, IRM enables finding invariant correlations that are robust to the changes in data distribution.
However, because of its challenging bi-level optimization form, they instantiate the IRM objective into more practical version, with fixed linear classifier constraint called IRMv1.
By assuming such classifier, they reformulate the hard constraints in IRM into the invariance penalty term, and in return, obtained reasonable OOD generalization performance.

However, despite the effectiveness of IRMv1 for OOD generalization, it lacks some formal guarantees which ideal IRM objective can provide~\cite{kamath2021does}.
IRMv1 and its variants inevitably fail when the number of training environments is smaller than the number of the spurious correlations existing in data~\cite{rosenfeld2020risks}.
In addition, we find that OOD generalization performance of IRMv1 and its variant severely degrades when the training data in each training environment become too small.

In this paper, we propose a novel approach for IRM objective initiation which learns invariant optimal predictors with meta-learning based optimization.
In this idea, instead of simplifying the challenging bi-level optimization objective of IRM, we keep the ideal objective form and solve it with Model-Agnostic Meta-Learning framework.
Specifically, as in the conventional meta-learning approach, we treat each training environment as one of tasks in meta-learning. 
But, we apply disjoint training environments for each inner loop  and outer loop (meta-loss) optimization, to introduce an invariance constraint to the classifier across different environments.

The main contribution of our work is providing a more effective initiation method of ideal IRM objective. 
While lifting the linear classifier assumption of IRMv1, our model still can find invariant representations existing across environments. 
We empirically show that the limitations of IRMv1 objective severely degrades OOD generalization performance even under simple conditions on data environment. 
Furthermore, we introduce a new auxiliary loss for non-linear invariant classifier which minimizes the standard deviation of meta-losses from all environments.
In experiments, we show that our meta-IRM framework achieves better OOD generalization performance than IRMv1 and all other alternatives.
Also it consistently provides invariant optimal predictor even when IRMv1 fails to generalize OOD.

\section{Related Work}

%\subsection{Variants of Invariant Risk Minimization}
There are several works on practical variants of IRM~\cite{ahuja2020invariant, krueger2020out}.
Invariant Risk Minimization Games~\cite{ahuja2020invariant} expands IRM objective by adopting game theory that finds the Nash equilibrium of classifiers among training environments.
Risk Extrapolation (MM-REx, V-REx)~\cite{krueger2020out} achieves OOD generalization with two objectives: minimizing average of empirical risks and maximizing similarity of empirical risks across training environments.
As there is a trade-off between these two objective, MM-REx uses mini-max objective over training risks and V-REx uses the variance of empirical risks as a regularization loss.
Our method  has similarity to V-REx in terms of considering the variance of risks, but we regulate the meta-losses for each environment rather than empirical risks.
Also, V-REx algorithm is mainly build upon regularization term to achieve OOD generalization when considering its large regularization coefficient.
In contrast, we normally assign very small coefficient for the standard deviation of meta-losses since our algorithm mainly focuses on bi-level optimization with meta-learning approach.
There are many research for IRM to understand its effect on OOD generalization~\cite{rosenfeld2020risks, ahuja2020empirical, choe2020empirical, kamath2021does}.
Analysis of \citet{rosenfeld2020risks} demonstrates that IRMv1 and its alternatives are failed to generalize OOD not only in the linear regime but also in the non-linear case.
We empirically show the limitation of IRMv1 in the linear regime and success of our meta-IRM in the same condition.
In addition, there is no explicit assumption of the linear classifier in our proposed method, unlike IRMv1.
Therefore, we expect our algorithm is able to provide OOD generalization even in nonlinear regimes.

%\subsection{Meta-Learning}
Meta-learning~\cite{schmidhuber1987evolutionary, bengio1990learning, thrun1998learning}, also known as learning-to-learn, is a learning framework devised for fast adaptation for unseen tasks. While training with multiple tasks, it learns quick knowledge or meta-knowledge~\cite{ravi2016optimization} and applies the knowledge to a new task adaptation for enhanced generalization performance.
Gradient-based meta-learning approaches~\cite{ravi2016optimization, finn2017model, nichol2018first} are widely researched methods and MAML~\cite{finn2017model} is one of such fundamental methods.
MAML consists of two levels of optimization, inner and outer.
Inner optimization updates parameters of the model adapting to a new task while outer or meta-optimization learns parameters which are needed for all other training tasks.
Our method is built upon this two level optimization framework where training tasks corresponding to training environments. 
However, our algorithm adopts a new learning strategy which is applying different training environments for inner and meta-optimization loop to learn invariant features existing across different environments.

There are several meta-learning approaches targeting the OOD problem.
However, their definition for OOD and the main purpose are different from ours.
\citet{jeong2020ood} introduce OOD-MAML, an extension of MAML, to detect unseen classes by learning artificially generated fake samples which are considered to be OOD samples.
They define OOD as samples from unseen classes and this definition is different from our goal of learning invariant correlations.
\citet{lee2019learning} suggest Bayesian-TAML for imbalanced and OOD tasks.
In Bayesian-TAML, the OOD generalization means relocating the initial parameters $\theta$ for adaptation to unseen OOD tasks, because meta-optimized $\theta$ may be less useful for the adaptation of OOD tasks.
In our approach, we do not rely on inner optimization or adaptation for the test environment, because we are aimed to find $\theta$ that can achieve the best OOD generalization by itself.

%\subsection{Bi-level Optimization}
For the bi-level optimization perspective, \citet{jenni2018deep} have a similar idea with our method. They reformulate the training procedure based on the principles of cross-validation to reduce overfitting and set up a bi-level optimization problem which uses splitted mini-batches as validation and training sets. 
They structured bi-level optimization by assigning each level with validation data optimization and the training data optimization.
In our algorithm, we also use different training environments for bi-level optimization, inner optimization and meta-optimization.
However, their work is intended to generalize for test data which have similar data distribution with training data, whereas we are targeting an invariance of predictor  even with the shift in data distribution.

\section{Proposed Method}
\subsection{Problem Set-Up}
We formalize our problem setting same as \citet{arjovsky2019invariant}.
We consider multiple $E$ training environments 
$\mathcal{E}_{tr}=\left\{e_i\right\}_{i=1}^E$ 
and datasets  $\left\{ (x_k^e, y_k^e) \right\}_{k=1}^n$ collected from each environment $e$.
The observational data $x_k^e$ include invariant feature set $\Psi_c$ and spurious feature set $\Psi_s$ in which both have strong correlation with target variable.
Each environment is related to one primary task but has a different data distribution.
The risk under environment $e$ with a predictor $f$ is defined as 
$R_e(f)= \mathbb{E}_{X^e, Y^e}\left[\ell(f(X^e), Y^e) \right]$,
where $\ell$ is a loss function, $f(X^e)$ is the output of the predictor and $Y^e$ is the target variable.
The purpose of IRM and our algorithm is to find a invariant predictor $f$ that minimizes the risk across unobserved but related environments $\mathcal{E}_{all}$ which includes $\mathcal{E}_{tr}$. In this setting, we expect our learned invariant causal predictor is supposed to be optimal across every related data environments $\mathcal{E}_{all}$.

\subsection{Invariant Risk Minimization}
Invariant Risk Minimization is intended to estimate optimal invariant causal predictor $f$ by learning correlations invariant across all of the training environments.
This also means to find a feature representation of data such that the optimal classifier is invariant across all environments.
Based on this idea, IRM is formulated as the following constrained bi-level optimization problem:

\begin{flalign}
    \begin{aligned}
        &\mathop{\min_{\Phi:\mathcal{X \rightarrow H}}}_{\omega:\mathcal{H \rightarrow Y}}\; \sum_{e\in\mathcal{E}_{\mathrm{tr}}} R^e (\omega \circ \Phi) \\
        &\text{subject to   } \omega \in \mathop{\text{arg}\min}_{\bar{\omega}:\mathcal{H \rightarrow Y}} R^e (\bar{\omega} \circ \Phi),
        \forall e \in \mathcal{E}_\mathrm{tr}
    \end{aligned}
\end{flalign}

where $\omega$ is a classifier (or final layer for regression problems), $\Phi$ is a data representation and $\omega \circ \Phi$ is a predictor.
However, it is difficult to solve such optimization problem directly since multiple constraints must be solved jointly.
\citet{arjovsky2019invariant} introduce IRMv1, a practical version of IRM, with Lagrangian form to relax joint constraints such that the classifier $\omega$ is "approximately locally optimal".
IRMv1 also assumes linear classifier $\omega$ as fixed scalar.
The learning objective of IRMv1 is as follows:

\begin{align}
    \min_{\Phi:\mathcal{X \rightarrow H}} \sum_{e\in\mathcal{E}_{\mathrm{tr}}} R^e (\Phi) + \lambda \cdot ||\nabla_{\omega|\omega=1.0} R^e(\omega \cdot \Phi) ||^2
\end{align}
where $\lambda \in (0,\infty]$ controls the invariance of the predictor $1 \cdot \Phi(x)$.
The gradient norm penalty $||\nabla_{\omega|\omega=1.0} R^e(\omega \cdot \Phi) ||^2$ indicates the optimality of fixed linear $\omega=1.0$.

Compared to an ideal IRM objective which possibly regards $\Psi_c$ with sampling noise as non-invariant features,
IRMv1 with fixed $\lambda$ is more robust for the sampling noise of finite samples~\cite{kamath2021does}.
However, its translated form also introduces severe limitations on the capabilities of ideal IRM objective. The linear fixed classifier assumption of IRMv1 can guarantee to find optimal invariant classifier under the condition that the number of spurious correlations is smaller than the number of training environment, $E > \left\vert \Psi_s \right\vert$ ~\cite{rosenfeld2020risks}.
With simple violation of this condition, $E \le \left\vert \Psi_s \right\vert$, IRMv1 can catastrophically fail to generalize for OOD. Furthermore, as the number of finite sample decreases, it is more likely to choose a fake invariant predictor which relies on environmental features, $\Psi_s$.~\cite{rosenfeld2020risks}.
In addition, we find that IRMv1 does not appropriately approximate IRM, even when  $E > \left\vert \Psi_s \right\vert$, if the training data from each environment are not sufficient. Despite its penalty term $||\nabla_{\omega|\omega=1.0} R^e(\omega \cdot \Phi) ||^2$ is close to zero, IRMv1 still can not find an invariant predictions, as shown in Figure~\ref{penalty}.

\begin{figure}[ht]
\vskip 0.2in
\begin{center}
\centerline{\includegraphics[width=\columnwidth]{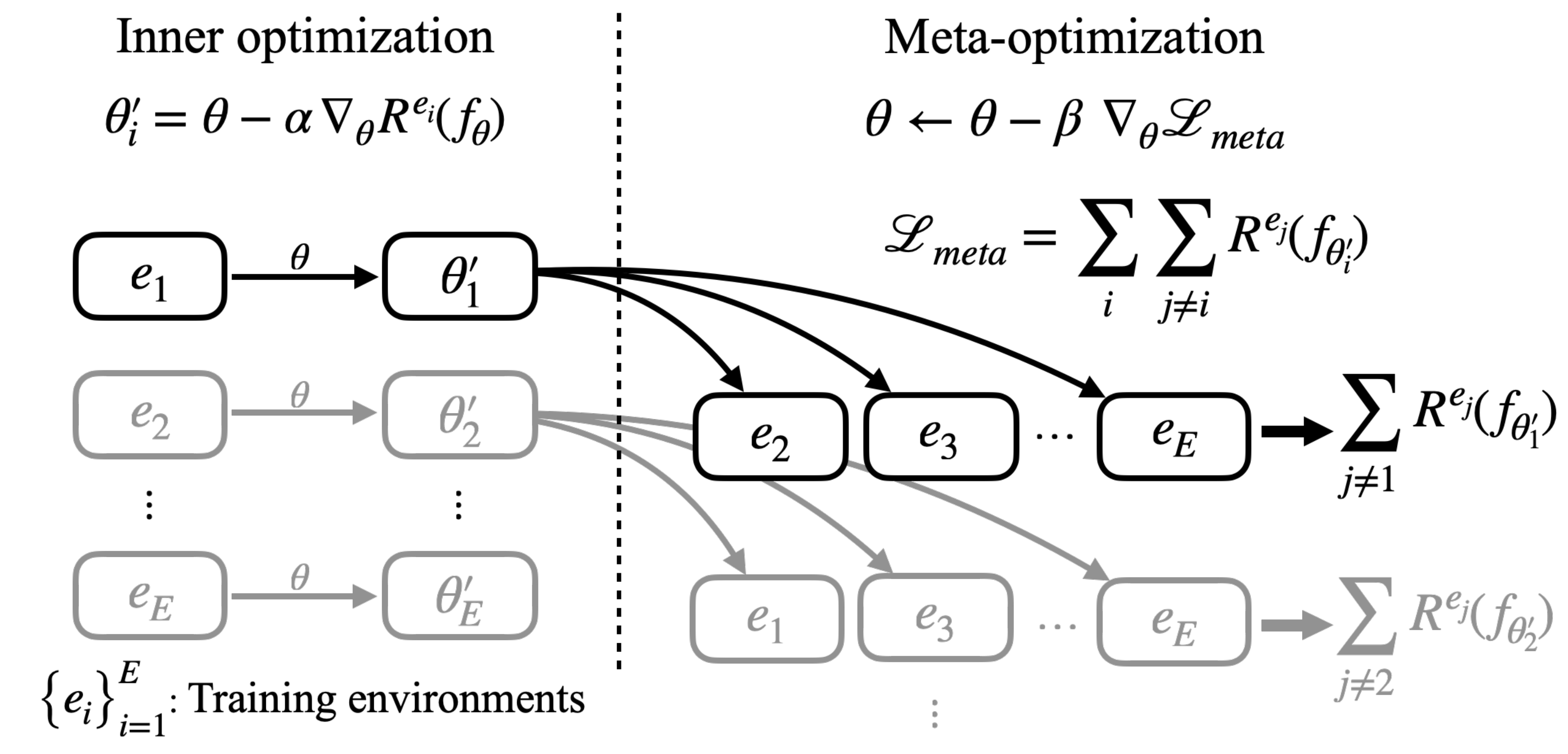}}
\caption{Schematic of learning process of meta-IRM. The model parameters $\theta$ are adapted to the training environments in the inner optimization. Each adapted parameter computes meta-loss from different environments.}
\label{scheamtic.png}
\end{center}
\vskip -0.3in
\end{figure}

\subsection{Meta-learned Invariant Risk Minimization}
We propose a novel learning framework for ideal IRM, meta-learned Invariant Risk Minimization (\textbf{meta-IRM}), which learns invariant optimal predictors with Model-Agnostic Meta-Learning framework.
Contrary to IRMv1, we keep the ideal bi-level formulation of IRM and do not impose
any linearity assumptions to the classifier $\omega$.
Instead, our meta-IRM jointly optimizes inner and upper-level objectives with the  meta-learning based optimization framework.
Conventional meta-learning approaches are intended to find proper initial parameters for fast adaptation on new tasks.
Our meta-IRM also finds optimal parameters that generalize well across not only for training environments but also for multiple unseen environments.
To estimate optimal predictor, meta-IRM performs inner optimization to each training environment and utilizes the gradients of meta-losses derived from all other training environments except the one used for inner optimization. 
As shown in Figure~\ref{grad.png}, while ERM supposed to converge to the $\theta_{ERM}$ by learning spurious correlations, meta-IRM merges gradient obtained from $f_{\theta'_1}$ and $f_{\theta'_2}$, to reach $\theta_{optimal}$.
Each gradient of meta-loss contributes to the converging path to $\theta_{optimal}$.

We consider multiple training tasks of meta-learning to be training environments of the IRM framework.
The predictor $\omega \circ \Phi$ is modeled as non-linear function $f_\theta$ with parameters $\theta$ and implemented as a neural network in our formulation.
Despite the form of the predictor is represented  as $f_\theta$ in our formulation, meta-IRM still can maintain the formulation of ideal IRM because of MAML.
MAML updates the final layer during the inner optimization and learns universal data representations while meta-optimization~\cite{raghu2019rapid}.
Similarly, in meta-IRM, the predictor $f_\theta$ is updated equivalently with the bi-level objective of IRM, optimizing classifier in the inner-level and learning feature representation in the upper-level.
For the inner-level optimization, the parameters $\theta$ of the predictor become $\theta_i'$ while adapting to the training environment $e_i \in \mathcal{E}_{tr}$.
This correspond to the inner optimization of MAML, except that there are multiple training environments for one task.
In formal expression, the inner optimization with one step gradient update can be defined as follows:
\begin{align}
    \theta_i' = \theta - \alpha \nabla_\theta R^{e_i}(f_\theta )
\end{align}
where $\alpha$ is a learning rate of the inner optimization.
We only consider one gradient update of inner loop update for the simplicity of notation and it is possible to use multiple gradient updates generally.

With inner updated predictor $f_{\theta_i'}$, meta-optimization across training environments with stochastic gradient descent(SGD) can be expressed as follows:
\begin{align}
\label{equation4}
    \theta \leftarrow \theta - \beta \ \nabla_\theta \sum_i \sum_{j} R^{e_j}(f_{\theta_{i}'}), \; e_j \sim \mathcal{E}_{tr} \setminus e_i
\end{align}

where $\beta$ is the learning rate of the meta-optimization, $e_j$ is training environment for meta-optimization for the parameters $\theta_i$ and $R^{e_j}(f_{\theta_{i}'})$ is the risk for meta-optimization, or meta-loss.
Note that the model $f_{\theta_i'}$ is evaluated with $e_j$ sampled from $\mathcal{E}_{tr} \setminus e_i$ in the meta-optimization, whereas it is updated with $e_i$ in the inner optimization, as we want to use the information obtained from the discrepancy among different training environments.
In addition, the meta-optimization is performed with the derivatives of $\theta_i'= \theta - \alpha \nabla_\theta R^{e_i}(f_\theta )$ through $\theta$ for all $i$, i.e., second-order derivatives of $\theta$.
The first-order approximation of meta-IRM is also applicable, but we show that it degrades the performance, see Section~\ref{section_ablation}.
The schematic of learning process of meta-IRM is shown in Figure~\ref{scheamtic.png}.

\begin{algorithm}[tb]
   \caption{Meta-IRM}
   \label{alg:example}
\begin{algorithmic}
   \STATE {\bfseries Require:} $\mathcal{E}_{tr}$: training environments
   \STATE {\bfseries Require:} $\mathcal{E}_{val}$: validation environments divided from $\mathcal{E}_{tr}$
   \STATE {\bfseries Initialize:} Randomly initialize $\theta$
   \WHILE {not done}
   \FOR {all $e_i \in \mathcal{E}_{tr}$}
   \STATE Evaluate $\nabla_\theta R^{e_i}(f_\theta )$ using $\left\{ (x_k^{e_i}, y_k^{e_i}) \right\}_{k=1}^n$
   \STATE Execute inner optimization: $\theta_i' = \theta - \alpha \nabla_\theta R^{e_i}(f_\theta )$
   \STATE Sample training environments $e_j$ from $\mathcal{E}_{tr} \setminus e_i$
   \FOR {all $e_j$}
   \STATE Compute meta-loss $R^{e_j}(f_{\theta_{i}'})$ using parameters $\theta_i$ and $\left\{ (x_k^{e_j}, y_k^{e_j}) \right\}_{k=1}^n$
   \ENDFOR
   \ENDFOR
   
   \STATE Compute standard deviation of $\left\{R^{e_j}(f_{\theta_{i}'})\right\}_{i,j}$
   \STATE Execute meta-optimization using Equation~\ref{equation5}
   \ENDWHILE

\end{algorithmic}
\end{algorithm}

We introduce auxiliary loss, the standard deviation of meta-losses in meta-optimization process, for the invariance of predictor.
The goal of IRM is learning invariant optimal predictor across environments.
With such auxiliary loss, meta-IRM updates the parameter $\theta$ to achieve invariant predictions across multiple adapted predictors $f_{\theta'_i}$ in meta-optimization.
In addition, the standard deviation loss makes meta-IRM enable more stable training and faster convergence.
In our experiment, by adding auxiliary loss, it is enough to achieve invariant predictions with using less inner update steps for meta-IRM.
Moreover, meta-IRM with auxiliary loss achieves the best results of OOD generalization in our experiment.
The auxiliary loss is defined as follows:
\begin{align}
\label{auxiliary loss}
    \sigma = \sqrt{\frac{1}{N}\sum _{k=1} ^N \left( \mathcal{L}^k_{meta} - \frac{1}{N}\sum_{l=1}^N\mathcal{L}^l_{meta} \right)^2}
\end{align}
where $N$ is the number of meta-losses and $\mathcal{L}^k_{meta}$ indicates the $k$-th meta-loss.
Applying the auxiliary loss, our meta-optimization update rule is changed from Equation~\ref{equation4} into as follows:
\begin{flalign}
\label{equation5}
\begin{aligned}
    \theta \leftarrow \theta - \beta \ \nabla_\theta \left\{  \sum_i \sum_j R^{e_j}(f_{\theta_{i}'})+\lambda \  \sigma \right\} , \; e_j \sim \mathcal{E}_{tr} \setminus e_i
\end{aligned}
\end{flalign}
where $\lambda$ is a hyperparameter enforcing meta-losses to have similar values and $\sigma$ is a standard deviation of meta-losses.
Without the auxiliary loss, meta-IRM still can achieve OOD generalization.
However, it is important to use such regularizing loss in order to stabilize the meta-IRM performance and provide invariant optimal predictor, see the ablation study in Section~\ref{section_ablation}.
The full algorithm of meta-IRM is outlined in Algorithm \ref{alg:example}.

In meta-learning process, meta-overfitting problem~\cite{zintgraf2019fast, lee2019meta} can occur when there are not enough training tasks (training environments in our case).
To address this problem, Gaussian DropGrad~\cite{tseng2020regularizing} is applied for meta-IRM with $N(1,\frac{p}{1-p})$ where $p$ is DropGrad rate.
We also adopt early stopping criterion while training meta-IRM to obtain optimal generalization performance of meta-learning method.
Similar to general machine learning algorithms, we consider validation loss of training environments as a proxy for the generalization error. 
However, we modified our early stopping criteria so that it can consider invariance of predictor. Our criteria stops model training if the standard deviation of validation losses from training environments decreases below the predefined threshold.
The main reason for using this stopping criterion is that lower validation error cannot always guarantee the invariant predictor, since  learning spurious correlations also can reduce validation loss.

\subsection{Connection between IRM and meta-IRM}
In this section, we illustrate how ideal IRM formulation is translated and initiated with meta-IRM approach.
In the formulation of IRMv1, each constraint of IRM is translated to squared norm of the gradients of the risk with respect to the fixed linear $\omega$, for the simplification of problem.
In our meta-IRM, which is built upon MAML, the challenging bi-level optimization of IRM is translated into two-level optimization scheme of meta-learning framework.
In this framework, the inner loop optimization corresponds to 
$\omega \in \mathop{\text{arg}\min}_{\bar{\omega}:\mathcal{H \rightarrow Y}} R^e (\bar{\omega} \circ \Phi)$
of IRM which updates the parameters $\theta$ to $\theta_i'$ for each training environment.
Meta-optimization is regarded as the upper-level optimization of IRM which updates the model parameters $\theta$ by minimizing sum of the meta-losses computed from $\theta'_i$ for all $i$.
As a consequence, the two-level optimization routines of meta-IRM solves challenging bi-level optimization of IRM without simplification, and it is much closer to the original IRM objective than IRMv1.
Furthermore, meta-IRM is also more robust to sampling noise of empirical distributions than IRMv1 in our experiment, see Figure~\ref{gap}.

Contrast to IRM, meta-IRM does not explicitly separate the data representation $\Phi$ and the classifier $\omega$ from predictor $f$ for optimization.
However, meta-IRM naturally follows IRM optimization structure, from the perspective of the classifier and data representation. 
As in the formulation of IRM, meta-IRM updates classifier in the inner loop optimization and learns data representation in the meta-optimization due to the property of MAML~\cite{raghu2019rapid}.
%Additionally, we apply auxiliary loss for the meta-optimization, which is designed to facilitate learning invariant predictors throughout the environments.

\section{Experiments and Results}

\subsection{Colored MNIST}
We evaluate meta-IRM with IRMv1 and its variants on Colored MNIST dataset from \citet{arjovsky2019invariant}.
It is modified MNIST images that are colored in either green or red.
These colors have spurious correlations with the class labels which should be decided by the shape of each digit.
Specifically, the digits from 0 to 4 have label $\Tilde{y}=0$ and 5 to 9 have label $\Tilde{y}=1$.
We add noise to $\Tilde{y}$ to build final label $y$ by flipping $\Tilde{y}$ with probability $\eta_e=0.25$.
To make spurious correlation, sample the color id $z$ by flipping $y$ with probability $p_e$ varying across environments.
Finally, color the digit with constructed color id as green if $z=0$ or red if $z=1$.
There are two training environments and one test environment.
We build validation set for early stopping by sampling from each training environments.
For the training environments, we use $p_e=0.1$ for the first environment and $p_e=0.2$ for the second environment.
The probability $p_e=0.9$ is used for the test environment where the correlation between color and label is reversed.
Each training environment has 25,000 images and test environment has 10,000 samples.
The goal of this task is to classify digits based on the shape of digits without considering colors.

\begin{table}[t]
\caption{Test accuracy (\%) of different algorithms on the Colored MNIST task in 10 trials (mean $\pm$ standard deviation). 
Test environments with $p_e=0.1$ and $p_e=0.2$ are similar to training environments. Test environment with $p_e=0.9$ has reversed direction of correlation between color and target label.}
\label{colored_mnist accuracies}
\vskip 0.15in
\begin{center}
\begin{small}
\begin{tabular}{lcccr}
\toprule
\multirow{2}{*}{\textbf{Algorithm}} & \multicolumn{3}{c}{\textbf{Test Accuracy}} \\
 & $p_e=0.1$ & $p_e=0.2$ & $p_e=0.9$ \\
\midrule
ERM & 88.6 $\pm$ 0.3 & 79.7 $\pm$ 0.6 & 16.4 $\pm$ 0.8 \\
IRMv1 & 71.4 $\pm$ 0.9 & 70.8 $\pm$ 1.0 & 66.9 $\pm$ 2.5 \\
MM-REx & 70.8 $\pm$ 1.5 & 70.4 $\pm$ 2.0 & 66.1 $\pm$ 4.9 \\
V-REx & 71.5 $\pm$ 0.8 & 71.1 $\pm$ 0.9 &  68.6  $\pm$ 1.2 \\
\textbf{meta-IRM (Ours)} & 70.9 $\pm$ 0.9 & 70.8 $\pm$ 1.0 & \textbf{70.4 $\pm$ 0.9} \\
\midrule
Random & 50 & 50 & 50 \\
ERM (grayscale) & 72.6 $\pm$ 0.3 & 72.7 $\pm$ 0.3 & 73.0 $\pm$ 0.5 \\
Optimal & 75  & 75 & 75 \\
\bottomrule
\end{tabular}
\end{small}
\end{center}
\vskip -0.2in
\end{table}

\begin{figure}[ht]
\vskip 0.2in
\begin{center}
\centerline{\includegraphics[width=\columnwidth]{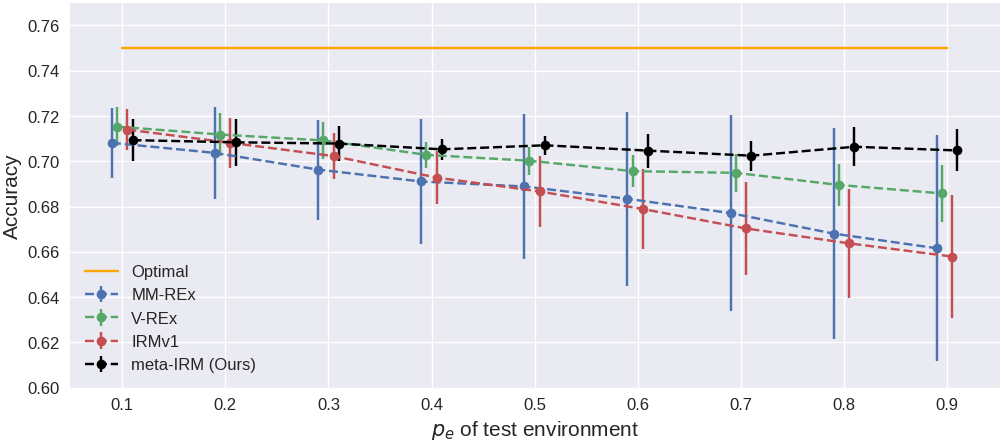}}
\caption{Accuracy of different training algorithms for the test environment with various $p_e$ on Colored MNIST. Dots indicate means and error bars represent standard deviations of 10 trials. We slightly shifted the dots to prevent error bars from overlapping.}
\label{various_models.png}
\end{center}
\vskip -0.3in
\end{figure}

We use MLP with ReLU activation function and a hidden layer to evaluate meta-IRM and others. 
Table~\ref{colored_mnist accuracies} shows the results of meta-IRM and other algorithms on test environments with $p_e$ as 0.1, 0.2 (similar to training environments), and 0.9 (out-of-distribution) on Colored MNIST task in 10 trials.
The ideal accuracy for test environment is 75\% due to the label noise $\eta_e=0.25$.
ERM achieves high test accuracy for for test environment which has similar distribution to training environments as $p_e=0.1$ and $p_e=0.2$ respectively.
However, it shows low accuracy for OOD worse than random selection.
In result, it predicts labels based on the color, which is a spurious feature, failing to generalize in a test environment with contradictory correlation between color and target label.
Our proposed meta-IRM achieves the highest mean accuracy of 70.4\% and the lowest standard deviation of 0.9\% compared with others.
ERM grayscale indicates ERM with grayscale MNIST images which do not include spurious features.
As shown in Figure~\ref{various_models.png}, we evaluated each algorithm by using test environment with diverse $p_e$ from 0.1 to 0.9 in steps of 0.1.
The predictions of meta-IRM are invariant across multiple test environments while others show worse performance as the environment moves away from training environments.

\subsection{Multi-Class Problem}
Many real-world problems are multi-class.
We compare meta-IRM with IRMv1 using 5-class and 10-class version of Colored MNIST~\cite{choe2020empirical}.
The number of colors is set to $k$ for $k$-class task and each color is assigned to a input channel of image.
For $k=5$, we consider two consecutive numbers as one class, such as 0-1, 2-3 etc.
For $k=10$, we consider each digit as one class.
Instead flipping labels with fixed noise $\eta_e$, we increase label index by 1 as noise.
For the label $\Tilde{y}=k$, the final label becomes $y=0$.

\begin{table}[t]
\caption{Train and test accuracy (\%) on multi-class ($k$=5, 10) Colored MNIST of different algorithms. ERM, IRMv1 and Grayscale results are cited from \citet{choe2020empirical}.}
\label{10class}
\vskip 0.15in
\begin{center}
\begin{small}
\begin{tabular}{lcccr}
\toprule
\multirow{2}{*}{\textbf{Algorithm}} & \multirow{2}{*}{\textbf{\# of class}} & \multicolumn{2}{c}{\textbf{Accuracy}} \\
                           &                              & Train            & Test            \\
\midrule
ERM                        & \multirow{6}{*}{5}            & 95.2 $\pm$   0.2           &     41.0 $\pm$ 0.6         \\
IRMv1                        &                              &  82.2 $\pm$ 0.4             &   62.0 $\pm$ 2.4           \\
\textbf{meta-IRM (Ours)} & &  76.4 $\pm$ 1.4& \textbf{74.0 $\pm$ 3.6} \\
Random                     &                              &        20       &        20      \\
ERM (grayscale)                  &                              &     73.2 $\pm$ 0.2          & 71.7$\pm$0.4             \\
Optimal & & 75 & 75 \\
\midrule
ERM                        & \multirow{6}{*}{10}            &      92.6 $\pm$ 0.2         &    39.2 $\pm$ 0.9          \\
IRMv1                        &                              &      83.4 $\pm$ 0.5         &    58.6 $\pm$ 2.5          \\
\textbf{meta-IRM (Ours)} & & 79.5 $\pm$ 0.6& \textbf{73.4 $ \pm$ 3.2} \\
Random                     &                              &     10          &       10       \\
ERM (grayscale)                  &                              &     73.2 $\pm$ 0.1          &   71.9 $\pm$ 0.5           \\
Optimal & & 75 & 75 \\
\bottomrule
\end{tabular}
\end{small}
\end{center}
\vskip -0.2in
\end{table}

The results of multi-class task is shown in Table~\ref{10class}.
Though the test environment results of IRMv1 still shows better achievement than ERM for $k=5$ and $k=10$, the performance degrades as the number of classes increases.
Also, IRMv1 absorbs small portion of spurious correlations as the train accuracy of IRMv1 has increased more than 75\%.
Our proposed meta-IRM achieves the best performance not only for $k=5$ but also for $k=10$ and shows invariant predictions across multiple environments despite having a slightly higher variance of test accuracy than $k=2$.

\begin{figure}[t]
\vskip 0.2in
\begin{center}
\centerline{\includegraphics[width=\columnwidth]{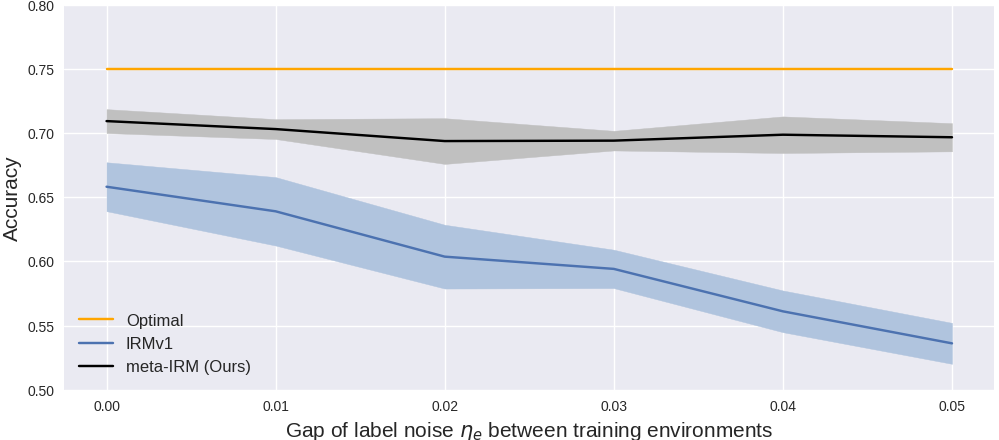}}
\caption{Accuracy of IRMv1 and meta-IRM for test environment with $\eta_e=0.25$, $p_e=0.9$ when the gap of $\eta_e$ between training environments increases.
The line indicates mean and colored area indicates a standard deviation of 10 trials.}
\label{gap}
\end{center}
\vskip -0.3in
\end{figure}

\subsection{Robustness to Sampling Noise}
We also experiment meta-IRM on Colored MNIST to evaluate its robustness to empirical distributions, even if it targets the ideal IRM objective.
In practice, finite samples cannot exactly represent the distribution, since they contain sampling noise.
\citet{kamath2021does} regard a small difference of $\eta_e$ between training environments as sampling noise.
We make a gap of $\eta_e$ between two training environments but still have the same average as 0.25~\cite{choe2020empirical}.
Training environment with $p_e=0.1$ has $\eta_e=0.25+\frac{gap}{2}$ and the other with $p_e=0.2$ has $\eta_e=0.25-\frac{gap}{2}$.
We increase the gap from 0 to 0.05 in steps of 0.01 and illustrate test environment accuracy of meta-IRM and IRMv1. 
As shown in Figure~\ref{gap}, meta-IRM shows robustness to noise while the performance of IRMv1 is getting worse as the gap increases.

\begin{figure}[t]
\vskip 0.2in
\begin{center}
\centerline{\includegraphics[width=\columnwidth]{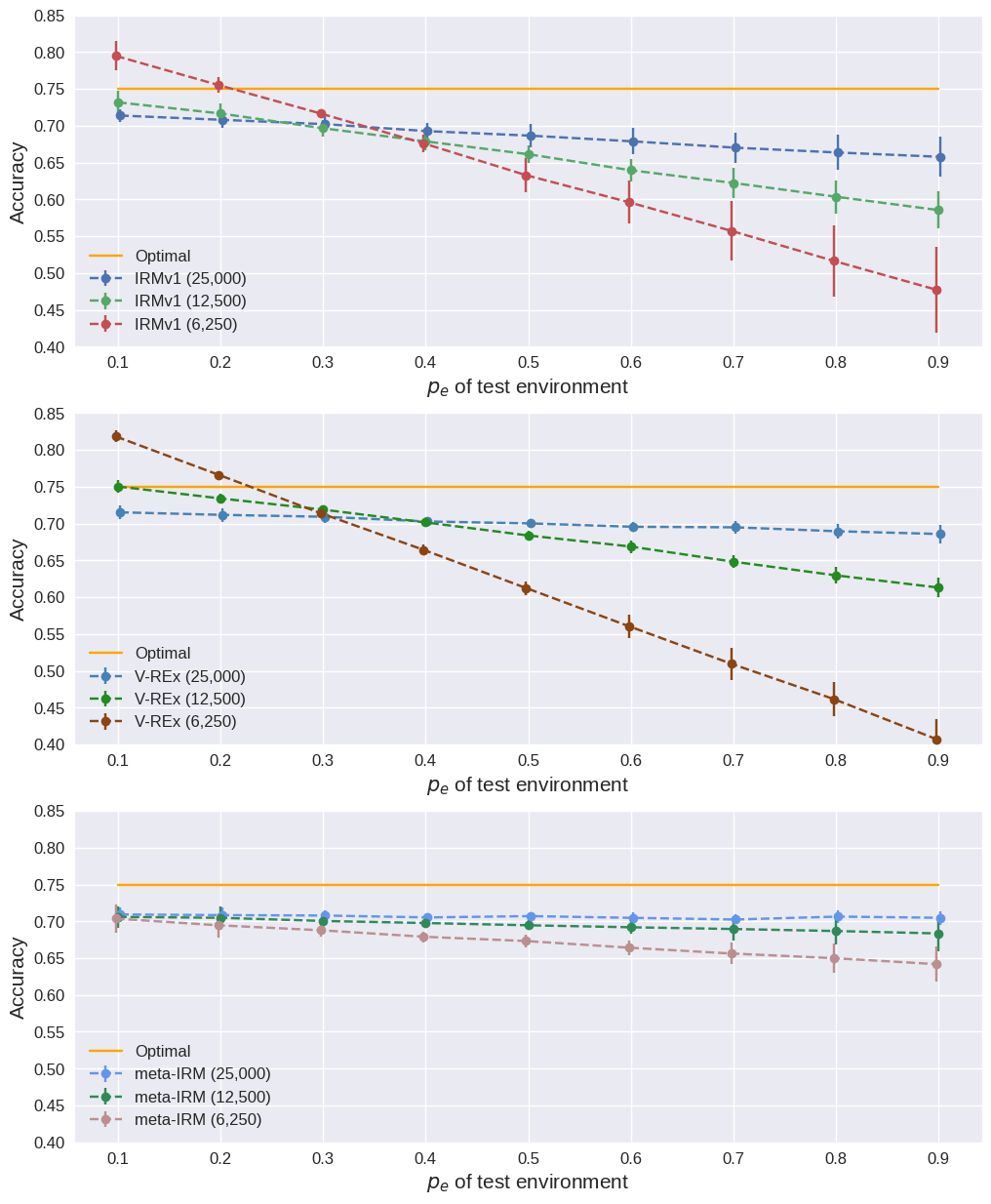}}
\caption{Accuracy of \textbf{Top: }IRMv1, \textbf{Middle: }V-REx, and \textbf{Bottom: }meta-IRM for the multiple test environments when training data samples are reduced from 25,000 to 12,500 and 6,250. Dots indicate means and error bars represent standard deviations of 10 trials. The number in parentheses indicates the number of training samples per training environment.}
\label{decreased_data}
\end{center}
\vskip -0.2in
\end{figure}

\subsection{The limitations of IRMv1}
\begin{figure}[t]
\vskip 0.2in
\begin{center}
\centerline{\includegraphics[width=\columnwidth]{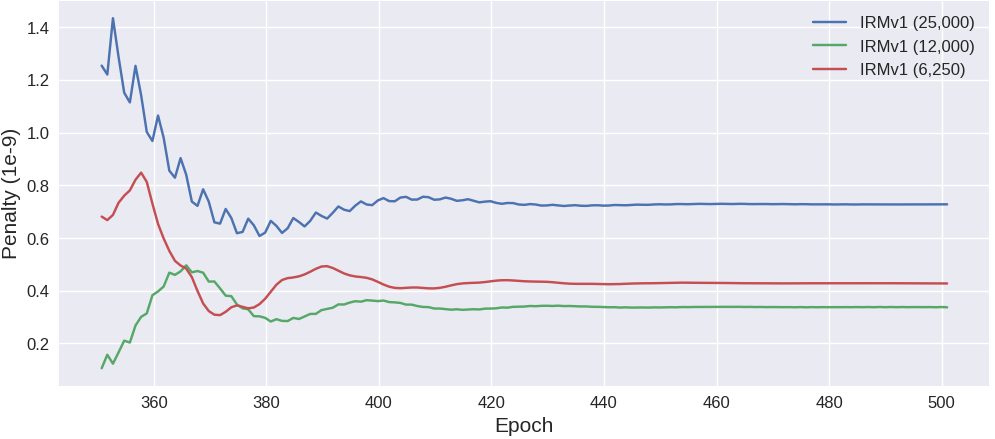}}
\caption{Penalty loss of IRMv1 when the number of training data samples per training environment is 25,000, 12,000, and 6,250.}
\label{penalty}
\end{center}
\vskip -0.2in
\end{figure}

From our experiments, we find that IRMv1 shows performance degradation with insufficient data in training environments.
We experiment with two cases: 12,500 and 6,250 training samples per each environment which become half and quarter, respectively.
IRMv1 shows worse test accuracy for OOD while meta-IRM shows comparable results as shown in Table~\ref{decreased_train_data}.
Figure~\ref{decreased_data} illustrates multiple test environments results of IRMv1 and meta-IRM in case of insufficient training data.
IRMv1 and V-REx start to follow ERM as the number of data decreases while meta-IRM still achieves invariant predictions in changing $p_e$ despite having insufficient training data.
As shown in Figure~\ref{penalty}, even if the penalty term of IRMv1 is close to zero regardless of the number of data samples, it fails to approximate ideal IRM objective.

\begin{table}[t]
\caption{Test accuracy (\%) of meta-IRM and IRMv1 on the Colored MNIST task when the number of training data per training environment is decreased (from 25,000 to 12,500). Results show (mean $\pm$ standard deviation) of 10 trials.}
\label{decreased_train_data}
\vskip 0.15in
\begin{center}
\begin{small}
\begin{tabular}{lcccr}
\toprule
\multirow{2}{*}{\textbf{Algorithm}} & \multicolumn{3}{c}{\textbf{Test Accuracy}} \\
 & $p_e=0.1$ & $p_e=0.2$ & $p_e=0.9$ \\
\midrule
IRMv1 & 73.2 $\pm$ 1.6 & 71.7 $\pm$ 1.3 & 58.5 $\pm$ 2.5 \\
V-REx & 75.0 $\pm$ 0.8 & 73.4 $\pm$ 0.6 & 61.3 $\pm$ 1.4 \\
\textbf{meta-IRM (Ours)} & 70.6 $\pm$ 1.5 & 70.5 $\pm$ 1.6 & \textbf{68.3 $\pm$ 2.3} \\
\midrule
Optimal & 75  & 75 & 75 \\
\bottomrule
\end{tabular}
\end{small}
\end{center}
\vskip -0.1in
\end{table}

In a real-world scenario, there are many kinds of spurious correlations in the data distribution.
However, IRMv1 can fail to achieve invariant predictor when the number of spurious correlations is greater than the number of training environments~\cite{rosenfeld2020risks}.
Creating lots of training environments can be a solution for such problem, but the number of training samples per training environment will be decreased since available data in real-world scenario is always limited.
Therefore, IRMv1 is going to fail to estimate invariant predictor in such condition.
In our experiments, we show that our meta-IRM performs invariant predictions not only for the case that the data is insufficient but also for many spurious correlations in the data.

\begin{figure}[t]
\vskip 0.2in
\begin{center}
\centerline{\includegraphics[width=\columnwidth]{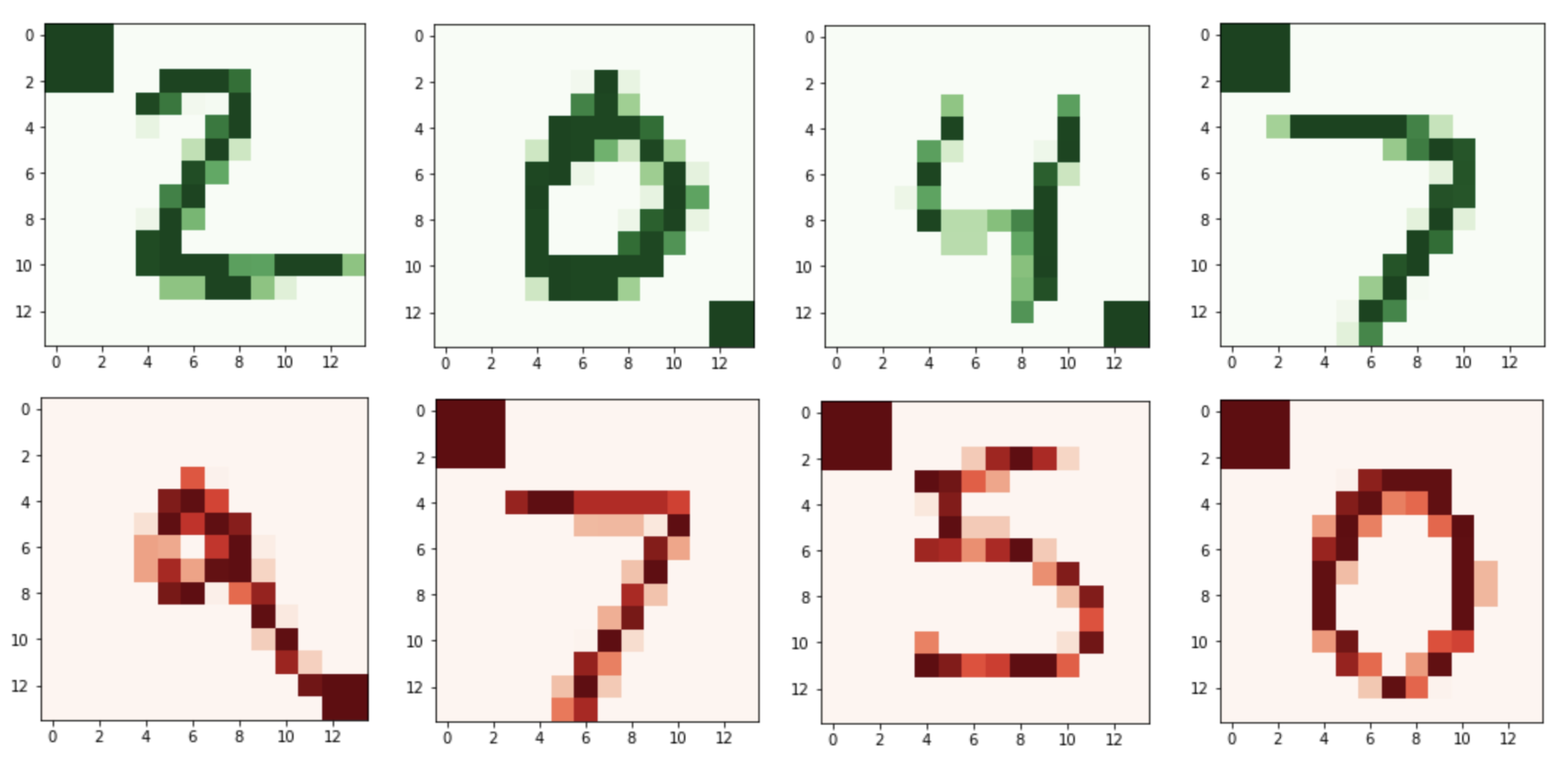}}
\caption{Image samples of Colored MNIST task with additional patch feature. These patches have strong but spuriouse correlation with target labels.}
\label{cm}
\end{center}
\vskip -0.2in
\end{figure}

\begin{figure}[t]
\vskip 0.2in
\begin{center}
\centerline{\includegraphics[width=\columnwidth]{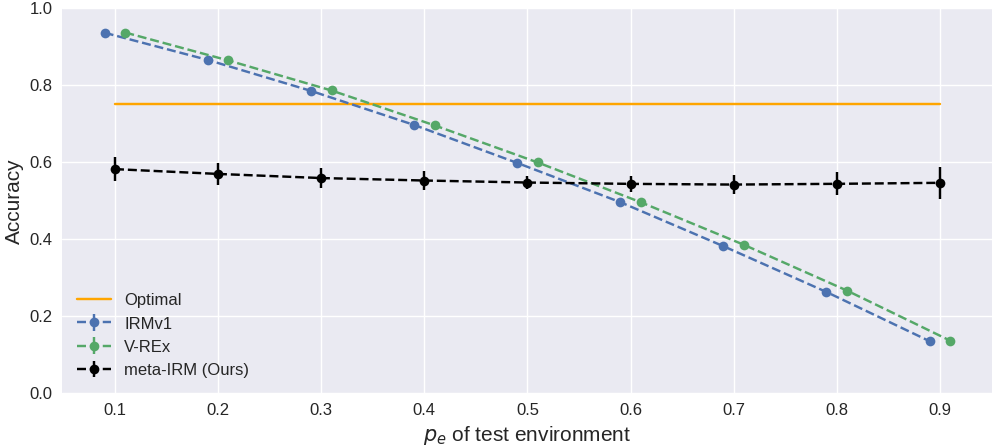}}
\caption{Test accuracy of different algorithms when the data include two spurious features, color and patch. Dots indicate means and error bars represent standard deviations of 10 trials.}
\label{2ss}
\end{center}
\vskip -0.4in
\end{figure}

We design additional spurious correlation as \cite{ahuja2020invariant}, by creating small patches of noise to the corner in the image where the locations of these patches are strongly correlated with the labels.
Similar with the probability $p_e$ of Colored MNIST, we define patch id $z'$ by flipping $y$ with probability $p_{e_2}$ also varying across environments.
Based on the patch id, we add $(3 \times 3)$ patch in the top left corner of the image if $z'=0$, or $(2 \times 2)$ patch in the bottom right corner of the image if  $z'=1$.
There are two training environments: $p_e=0.1$, $p_{e_2}=0.2$ for the first and $p_e=0.2$, $p_{e_2}=0.1$ for the second and one test environment with $p_e=0.9$ and $p_{e_2}=0.9$.
The patches are independent with the colors of digits but have strong correlation with colors and target labels.
Figure~\ref{cm} illustrates the image samples of Colored MNIST task with additional spurious features.
In this experiment, IRMv1 fails to generalize to OOD, following ERM.
IRMv1 achieves high test accuracy for the test environments analogous to the training environments but shows low accuracy for the test environments with $p_e=0.9$ and $p_{e_2}=0.9$.
Unlike IRMv1, meta-IRM achieves invariant predictions across multiple environments although the performance is slightly degraded as shown in see Table~\ref{two_sp}.
Figure~\ref{2ss} illustrates results of meta-IRM and IRMv1.
Since IRMv1 learns spurious features, the results of for test environments with $p_e=0.1$ and $0.2$ seem plausible.
However, its test accuracy decreases for OOD.

\begin{table}[t]
\caption{Test accuracy (\%) of meta-IRM and IRMv1 on the Colored MNIST task when there are two spurious features (color and patch) in the data. Results show (mean $\pm$ standard deviation) of 10 trials. The probability $p_{e_2}$ is set to the same as $p_e$.}
\label{two_sp}
\vskip 0.15in
\begin{center}
\begin{small}
\begin{tabular}{lcccr}
\toprule
\multirow{2}{*}{\textbf{Algorithm}} & \multicolumn{3}{c}{\textbf{Test Accuracy}} \\
 & $p_e=0.1$ & $p_e=0.2$ & $p_e=0.9$ \\
\midrule
IRMv1 & 93.5 $\pm$ 0.2 & 86.4 $\pm$ 0.3 & 13.4 $\pm$ 0.3 \\
V-REx & 93.6 $\pm$ 0.4 & 86.3 $\pm$ 0.3 & 13.5 $\pm$ 0.3 \\
\textbf{meta-IRM (Ours)} & 58.1 $\pm$ 3.1 & 56.8 $\pm$ 2.9 & \textbf{54.5 $\pm$ 4.0} \\
\midrule
Optimal & 75  & 75 & 75 \\
\bottomrule
\end{tabular}
\end{small}
\end{center}
\vskip -0.1in
\end{table}

\subsection{Ablation Study}
\label{section_ablation}
In this section, we show ablation experiments to demonstrate the importance of the standard deviation loss.
Also, meta-IRM algorithm uses second-order derivatives of $\theta$ and considers different training environments between inner optimization and meta-loss computation.
We perform ablation study of meta-IRM in order to understand the effects.
Without using the additional standard deviation loss, meta-IRM still shows comparable accuracy for OOD.
However, it is recommended to use such auxiliary loss to stabilize the algorithm and achieve the best performance.
We also show that it is important to use second-order derivatives of $\theta$ in the objective of meta-IRM to achieve high-performance of OOD generalization.
Furthermore, using different training environments for inner optimization and meta-loss computation is critical to learn invariant predictors as shown in Table~\ref{ablation}.

\begin{table}[t]
\caption{Ablation study result of using standard deviation of meta-losses as auxiliary loss on Colored MNIST task. }
\label{ablation}
\vskip 0.15in
\begin{center}
\begin{small}
\begin{tabular}{lcccr}
\toprule
\multirow{2}{*}{\textbf{Algorithm}} & \multicolumn{2}{c}{\textbf{Test Accuracy}} \\
 & $p_e=0.1$ & $p_e=0.9$ \\
\midrule
meta-IRM (w/o std. loss) & 73.0 $\pm$ 0.8 & 64.2 $\pm$ 3.2 \\
meta-IRM (first-order approx.) & 62.6 $\pm$ 5.5 & 59.1 $\pm$ 6.3 \\
meta-IRM (using same env.) & 89.3 $\pm$ 1.6 & 13.6 $\pm$ 6.6 \\
\textbf{meta-IRM (Ours)} & 70.9 $\pm$ 0.9 & \textbf{70.4 $\pm$ 0.9} \\
\midrule
Optimal & 75  & 75 \\
\bottomrule
\end{tabular}
\end{small}
\end{center}
\vskip -0.1in
\end{table}

\section{Conclusion}
In this work, we interpret IRM objective in meta-learning perspective and propose a novel meta-learning based method for ideal IRM initiation to obtain further enhanced OOD generalization.
Our method maintains the bi-level optimization form of ideal IRM objective by efficiently relaxing the linearity restriction of classifier based on MAML.
In experiments, meta-IRM shows superior OOD generalization performance than all other IRM variants and is able to provide invariant predictor even when IRMv1 and all its variants fail. 
Furthermore, our method shows more robustness to the noise in true causal features, although original IRM objective is susceptible to such condition.
Our meta-IRM provides the most effective IRM initiation method for OOD generalization and also shows promising results for the non-linear regime by excluding the linearity assumption for predictor.

% In the unusual situation where you want a paper to appear in the
% references without citing it in the main text, use \nocite

\bibliography{example_paper}

\begin{thebibliography}{28}
\providecommand{\natexlab}[1]{#1}
\providecommand{\url}[1]{\texttt{#1}}
\expandafter\ifx\csname urlstyle\endcsname\relax
  \providecommand{\doi}[1]{doi: #1}\else
  \providecommand{\doi}{doi: \begingroup \urlstyle{rm}\Url}\fi

\bibitem[Ahuja et~al.(2020{\natexlab{a}})Ahuja, Shanmugam, Varshney, and
  Dhurandhar]{ahuja2020invariant}
Ahuja, K., Shanmugam, K., Varshney, K., and Dhurandhar, A.
\newblock Invariant risk minimization games.
\newblock In \emph{International Conference on Machine Learning}, pp.\
  145--155. PMLR, 2020{\natexlab{a}}.

\bibitem[Ahuja et~al.(2020{\natexlab{b}})Ahuja, Wang, Dhurandhar, Shanmugam,
  and Varshney]{ahuja2020empirical}
Ahuja, K., Wang, J., Dhurandhar, A., Shanmugam, K., and Varshney, K.~R.
\newblock Empirical or invariant risk minimization? a sample complexity
  perspective.
\newblock \emph{arXiv preprint arXiv:2010.16412}, 2020{\natexlab{b}}.

\bibitem[Arjovsky et~al.(2019)Arjovsky, Bottou, Gulrajani, and
  Lopez-Paz]{arjovsky2019invariant}
Arjovsky, M., Bottou, L., Gulrajani, I., and Lopez-Paz, D.
\newblock Invariant risk minimization.
\newblock \emph{arXiv preprint arXiv:1907.02893}, 2019.

\bibitem[Beery et~al.(2018)Beery, Van~Horn, and Perona]{beery2018recognition}
Beery, S., Van~Horn, G., and Perona, P.
\newblock Recognition in terra incognita.
\newblock In \emph{Proceedings of the European Conference on Computer Vision
  (ECCV)}, pp.\  456--473, 2018.

\bibitem[Bengio et~al.(1990)Bengio, Bengio, and Cloutier]{bengio1990learning}
Bengio, Y., Bengio, S., and Cloutier, J.
\newblock \emph{Learning a synaptic learning rule}.
\newblock Citeseer, 1990.

\bibitem[Choe et~al.(2020)Choe, Ham, and Park]{choe2020empirical}
Choe, Y.~J., Ham, J., and Park, K.
\newblock An empirical study of invariant risk minimization.
\newblock \emph{arXiv preprint arXiv:2004.05007}, 2020.

\bibitem[de~Haan et~al.(2019)de~Haan, Jayaraman, and Levine]{de2019causal}
de~Haan, P., Jayaraman, D., and Levine, S.
\newblock Causal confusion in imitation learning.
\newblock In \emph{Advances in Neural Information Processing Systems}, pp.\
  11698--11709, 2019.

\bibitem[Finn et~al.(2017)Finn, Abbeel, and Levine]{finn2017model}
Finn, C., Abbeel, P., and Levine, S.
\newblock Model-agnostic meta-learning for fast adaptation of deep networks.
\newblock In \emph{International Conference on Machine Learning}, pp.\
  1126--1135. PMLR, 2017.

\bibitem[Geirhos et~al.(2018)Geirhos, Rubisch, Michaelis, Bethge, Wichmann, and
  Brendel]{geirhos2018imagenet}
Geirhos, R., Rubisch, P., Michaelis, C., Bethge, M., Wichmann, F.~A., and
  Brendel, W.
\newblock Imagenet-trained cnns are biased towards texture; increasing shape
  bias improves accuracy and robustness.
\newblock \emph{arXiv preprint arXiv:1811.12231}, 2018.

\bibitem[Heinze-Deml et~al.(2018)Heinze-Deml, Peters, and
  Meinshausen]{heinze2018invariant}
Heinze-Deml, C., Peters, J., and Meinshausen, N.
\newblock Invariant causal prediction for nonlinear models.
\newblock \emph{Journal of Causal Inference}, 6\penalty0 (2), 2018.

\bibitem[Ilyas et~al.(2019)Ilyas, Santurkar, Tsipras, Engstrom, Tran, and
  Madry]{ilyas2019adversarial}
Ilyas, A., Santurkar, S., Tsipras, D., Engstrom, L., Tran, B., and Madry, A.
\newblock Adversarial examples are not bugs, they are features.
\newblock In \emph{Advances in Neural Information Processing Systems}, pp.\
  125--136, 2019.

\bibitem[Jenni \& Favaro(2018)Jenni and Favaro]{jenni2018deep}
Jenni, S. and Favaro, P.
\newblock Deep bilevel learning.
\newblock In \emph{Proceedings of the European conference on computer vision
  (ECCV)}, pp.\  618--633, 2018.

\bibitem[Jeong \& Kim(2020)Jeong and Kim]{jeong2020ood}
Jeong, T. and Kim, H.
\newblock Ood-maml: Meta-learning for few-shot out-of-distribution detection
  and classification.
\newblock \emph{Advances in Neural Information Processing Systems}, 33, 2020.

\bibitem[Kamath et~al.(2021)Kamath, Tangella, Sutherland, and
  Srebro]{kamath2021does}
Kamath, P., Tangella, A., Sutherland, D.~J., and Srebro, N.
\newblock Does invariant risk minimization capture invariance?
\newblock \emph{arXiv preprint arXiv:2101.01134}, 2021.

\bibitem[Krueger et~al.(2020)Krueger, Caballero, Jacobsen, Zhang, Binas, Priol,
  and Courville]{krueger2020out}
Krueger, D., Caballero, E., Jacobsen, J.-H., Zhang, A., Binas, J., Priol,
  R.~L., and Courville, A.
\newblock Out-of-distribution generalization via risk extrapolation (rex).
\newblock \emph{arXiv preprint arXiv:2003.00688}, 2020.

\bibitem[Lee et~al.(2019{\natexlab{a}})Lee, Lee, Na, Kim, Park, Yang, and
  Hwang]{lee2019learning}
Lee, H.~B., Lee, H., Na, D., Kim, S., Park, M., Yang, E., and Hwang, S.~J.
\newblock Learning to balance: Bayesian meta-learning for imbalanced and
  out-of-distribution tasks.
\newblock \emph{arXiv preprint arXiv:1905.12917}, 2019{\natexlab{a}}.

\bibitem[Lee et~al.(2019{\natexlab{b}})Lee, Maji, Ravichandran, and
  Soatto]{lee2019meta}
Lee, K., Maji, S., Ravichandran, A., and Soatto, S.
\newblock Meta-learning with differentiable convex optimization.
\newblock In \emph{Proceedings of the IEEE/CVF Conference on Computer Vision
  and Pattern Recognition}, pp.\  10657--10665, 2019{\natexlab{b}}.

\bibitem[Morcos et~al.(2018)Morcos, Raghu, and Bengio]{morcos2018insights}
Morcos, A.~S., Raghu, M., and Bengio, S.
\newblock Insights on representational similarity in neural networks with
  canonical correlation.
\newblock \emph{arXiv preprint arXiv:1806.05759}, 2018.

\bibitem[Nichol et~al.(2018)Nichol, Achiam, and Schulman]{nichol2018first}
Nichol, A., Achiam, J., and Schulman, J.
\newblock On first-order meta-learning algorithms.
\newblock \emph{arXiv preprint arXiv:1803.02999}, 2018.

\bibitem[Pennington et~al.(2014)Pennington, Socher, and
  Manning]{pennington2014glove}
Pennington, J., Socher, R., and Manning, C.~D.
\newblock Glove: Global vectors for word representation.
\newblock In \emph{Proceedings of the 2014 conference on empirical methods in
  natural language processing (EMNLP)}, pp.\  1532--1543, 2014.

\bibitem[Peters et~al.(2016)Peters, B{\"u}hlmann, and
  Meinshausen]{peters2016causal}
Peters, J., B{\"u}hlmann, P., and Meinshausen, N.
\newblock Causal inference by using invariant prediction: identification and
  confidence intervals.
\newblock \emph{Journal of the Royal Statistical Society. Series B (Statistical
  Methodology)}, pp.\  947--1012, 2016.

\bibitem[Raghu et~al.(2019)Raghu, Raghu, Bengio, and Vinyals]{raghu2019rapid}
Raghu, A., Raghu, M., Bengio, S., and Vinyals, O.
\newblock Rapid learning or feature reuse? towards understanding the
  effectiveness of maml.
\newblock \emph{arXiv preprint arXiv:1909.09157}, 2019.

\bibitem[Ravi \& Larochelle(2016)Ravi and Larochelle]{ravi2016optimization}
Ravi, S. and Larochelle, H.
\newblock Optimization as a model for few-shot learning.
\newblock 2016.

\bibitem[Rosenfeld et~al.(2020)Rosenfeld, Ravikumar, and
  Risteski]{rosenfeld2020risks}
Rosenfeld, E., Ravikumar, P., and Risteski, A.
\newblock The risks of invariant risk minimization.
\newblock \emph{arXiv preprint arXiv:2010.05761}, 2020.

\bibitem[Schmidhuber(1987)]{schmidhuber1987evolutionary}
Schmidhuber, J.
\newblock \emph{Evolutionary principles in self-referential learning, or on
  learning how to learn: the meta-meta-... hook}.
\newblock PhD thesis, Technische Universit{\"a}t M{\"u}nchen, 1987.

\bibitem[Thrun \& Pratt(1998)Thrun and Pratt]{thrun1998learning}
Thrun, S. and Pratt, L.
\newblock Learning to learn: Introduction and overview.
\newblock In \emph{Learning to learn}, pp.\  3--17. Springer, 1998.

\bibitem[Tseng et~al.(2020)Tseng, Chen, Tsai, Liu, Lin, and
  Yang]{tseng2020regularizing}
Tseng, H.-Y., Chen, Y.-W., Tsai, Y.-H., Liu, S., Lin, Y.-Y., and Yang, M.-H.
\newblock Regularizing meta-learning via gradient dropout.
\newblock In \emph{Proceedings of the Asian Conference on Computer Vision},
  2020.

\bibitem[Zintgraf et~al.(2019)Zintgraf, Shiarli, Kurin, Hofmann, and
  Whiteson]{zintgraf2019fast}
Zintgraf, L., Shiarli, K., Kurin, V., Hofmann, K., and Whiteson, S.
\newblock Fast context adaptation via meta-learning.
\newblock In \emph{International Conference on Machine Learning}, pp.\
  7693--7702. PMLR, 2019.

\end{thebibliography}

\newpage
\phantom{blahblah}
\newpage
\appendix

\section{Additional Results}
\subsection{Representational Similarity Experiment}
\begin{figure}[h]
\vskip 0.1in
\begin{center}
\centerline{\includegraphics[width=0.9\columnwidth]{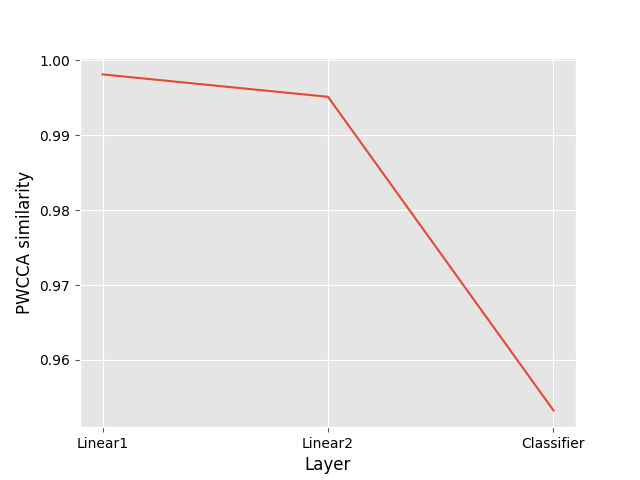}}
\caption{PWCCA similarity between representations of each layer before and after inner optimization.}
\label{pwcca}
\end{center}
\vskip -0.4in
\end{figure}

We show optimization characteristic of meta-IRM by measuring similarity between feature representations of each layer in the predictor before and after inner optimization~\cite{raghu2019rapid}.
We apply Projected Weighted Canonical Correlation Analysis (PWCCA) similarity~\cite{morcos2018insights} to the feature representations.
We experiment meta-IRM on Colored MNIST without using auxiliary loss in order to understand the effect of two-level optimization of meta-learning.
We illustrate PWCCA similarities of representations just before the training is converged.
As shown in Figure~\ref{pwcca}, feature representations of the first and second layers are highly similar, however, the classifier has a relatively low PWCCA similarity.
The result demonstrates that the inner optimization mostly updates the classifier whereas the data representation has little functional changes.
Consequently, meta-IRM can learn invariant optimal predictor under the original bi-level optimization of IRM objective.

\subsection{Invariant Data Representation}
\begin{figure}[t]
\vskip 0.2in
\begin{center}
\centerline{\includegraphics[width=\columnwidth]{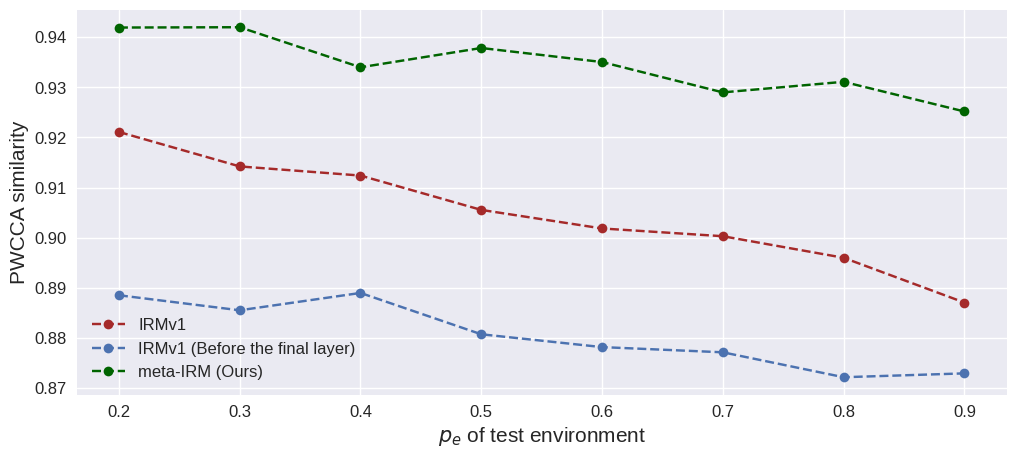}}
\caption{PWCCA similarity of data representations on the test environment between $p_e=0.1$ and multiple $p_e$ values.}
\label{data_rep}
\end{center}
\vskip -0.4in
\end{figure}

We already show the invariant predictions of meta-IRM as shown in Figure~\ref{various_models.png}.
The data representation which learns invariant correlations is invariant across environments.
We experiment invariance of the data representation of IRMv1 and meta-IRM by using PWCCA similarity.
The data representation of the predictor trained by each algorithm is obtained from the test environments with diverse $p_e$ from 0.1 to 0.9 in steps of 0.1.
We consider the data representation of the case of meta-IRM to be feature vectors which are the input of the classifier layer.
For the case of IRMv1, the data representation is the output vector of the neural network due to its formulation.
We compute the PWCCA similarity between the data representation obtained from the test environment with $p_e=0.1$ and others.
Figure~\ref{data_rep} illustrates the results of the PWCCA similarity between the data representations across multiple test environments.
The similarity between data representations of IRMv1 decreases as $p_e$ of test environment increases.
In contrast, meta-IRM shows more invariant data representations across entire test environments with higher similarity scores than IRMv1.
Therefore, meta-IRM is better to learn invariant optimal predictor by learning invariant correlations across environments than IRMv1.

\begin{table}[t]
\caption{Test environment accuracy (\%) on PunctuatedSST-2 task of different algorithms. ERM (grayscale) indicates the results of ERM with vanilla SST-2 dataset.}
\label{punctuated}
\vskip 0.15in
\begin{center}
\begin{small}
\begin{tabular}{lcccr}
\toprule
\textbf{Algorithm} & $\eta_e$ & Test accuracy ($p_e=0.9$) \\
\midrule
ERM                        & \multirow{5}{*}{0.25}        &     30.7 $\pm$ 1.5         \\
IRMv1                        &                         &   62.0 $\pm$ 1.9           \\
\textbf{meta-IRM (Ours)} & & \textbf{62.2 $\pm$ 1.8} \\
ERM (grayscale)                  &                                & 62.3 $\pm$ 0.5             \\
Optimal & & 75\\
\midrule
ERM                        & \multirow{5}{*}{0}             &    56.2 $\pm$ 2.9          \\
IRMv1                        &                             &    67.4 $\pm$ 1.4          \\
\textbf{meta-IRM (Ours)} & & \textbf{73.0 $\pm$ 0.7} \\
ERM (grayscale)                  &                                  &   76.7 $\pm$ 2.7           \\
Optimal & & 100 \\
\bottomrule
\end{tabular}
\end{small}
\end{center}
\vskip -0.1in
\end{table}

\section{PunctuatedSST-2}
We evaluate our meta-IRM on natural language processing (NLP) task using PunctuatedSST-2 dataset from \citet{choe2020empirical}.
PunctuatedSST-2 is a modified version of NLP benchmark dataset, Standford Sentiment Treebank (SST-2).
Similar to the spurious feature 'color' in Colored MNIST task, we artificially make spurious features using punctuation marks.
%We do not consider label noise $\eta_e$ since SST-2 task is already hard to solve with the baseline model, simple MLP with GloVe embedding.
We add noise to target labels to construct final label $y$ by flipping with probability $\eta_e$.
Punctuation mark id $z$ is constructed by flipping $y$ with probability $p_e$.
Finally, punctuate the input with an exclamation mark (!) if $z=0$ or with a period (.) if $z=1$.
There are two training environments with $p_e=0.1$ and $p_e=0.2$, respectively, and one test environment $p_e=0.9$.

We train the MLP with ReLU activation function and a hidden layer.
The input data are embedded with GloVe~\cite{pennington2014glove}.
As shown in Table~\ref{punctuated}, both IRMv1 and our meta-IRM show the results close to the optimal when the label noise $\eta_e=0.25$.
We also show the results when the label noise $\eta_e=0$.
Our meta-IRM shows much higher test accuracy than IRMv1 and ERM, close to ERM grayscale when $\eta_e=0$.

\bibliographystyle{icml2021}
\end{document}